\documentclass[11pt,letterpaper]{article}

\usepackage{template/eacl2017}

\def\arXiv{1}  

\if\arXiv1
\fi

\eaclfinalcopy 

\author{Jie Mei\textsuperscript{1},
        Aminul Islam\textsuperscript{2},
        Yajing Wu\textsuperscript{1},
        Abidalrahman Moh'd\textsuperscript{1},
        Evangelos E. Milios\textsuperscript{1}
        \vspace{0.5em}\\
  \textsuperscript{1} Faculty of Computer Science, Dalhousie University \\
  {\tt \{jmei, yajing, amohd, eem\}@cs.dal.ca}
  \vspace{0.3em}\\
  \textsuperscript{2} School of Computing and Informatics, University of Louisiana at Lafayette \\
  {\tt aminul@louisiana.edu} \\
  }

\date{}

\usepackage{amsmath}      
\usepackage{times}        
\usepackage{url}
\usepackage{graphicx}

\usepackage{booktabs}
\usepackage{pbox}
\usepackage{adjustbox}
\usepackage{upquote}
\usepackage{tabularx}
\usepackage{multirow}

\usepackage{array}
\newcolumntype{L}[1]{>{\raggedright\let\newline\\\arraybackslash\hspace{0pt}}m{#1}}
\newcolumntype{C}[1]{>{\centering\let\newline\\\arraybackslash\hspace{0pt}}m{#1}}
\newcolumntype{R}[1]{>{\raggedleft\let\newline\\\arraybackslash\hspace{0pt}}m{#1}}

\usepackage{pgfplots, pgfplotstable}
\usetikzlibrary{patterns}
\usepgfplotslibrary{groupplots, units}

\title{Statistical Learning for OCR Text Correction}

\begin{document}


\maketitle

\begin{abstract}
  The accuracy of Optical Character Recognition (OCR) is crucial to the success of subsequent applications used in text analyzing pipeline.
Recent models of OCR post-processing significantly improve the quality of OCR-generated text, but are still prone to suggest correction candidates from limited observations while insufficiently accounting for the characteristics of OCR errors.
In this paper, we show how to enlarge candidate suggestion space by using external corpus and integrating OCR-specific features in a regression approach to correct OCR-generated errors.
The evaluation results show that our model can correct 61.5\% of the OCR-errors (considering the top 1 suggestion) and 71.5\% of the OCR-errors (considering the top 3 suggestions), for cases where the theoretical correction upper-bound is 78\%.

\end{abstract}

\section{Introduction}
\label{sec:Introduction}



An increasing amount of data is produced and transformed into the digital form these days, including magazines, books, and scientific articles.
Using the graphic formats, like Portable Document Format (PDF) or Joint Picture Group (JPG), is a comprehensive solution for efficient digitization as well as better preserving the page layout and the graphical information (i.e., charts and figures).
Since information in such formats is not machine-readable, analyzing such data relies heavily on the accuracy of Optical Character Recognition (OCR)~\cite{Doermann:1998:IRD:292300.292304}.
However, OCR systems are imperfect and prone to errors.

Post-processing is an important step in improving the quality of OCR output, which is crucial to the success of any text analyzing system in pipeline.
An OCR Post-processing model attempts to detect misspellings in noisy OCR output and correct such errors to their intended representations.
Many machine learning approaches~\cite{Lund:2009:IOC:1555400.1555437,6065414,doi:10.1117/12.2006228,doi:10.1117/12.2042502} correct the OCR-generated errors by selecting the most appropriate correction among candidates.
OCR-generated errors are more diverse than handwriting errors in many aspects~\cite{Jones:1991:IMK,Kukich:1992:TAC:146370.146380}.
Machine learning approaches incorporate different features enabling more robust candidate selection, instead of inferring from limited observations, for example, using a probabilistic-based model~\cite{Taghva:2001:OCRSpell}.

While machine learning approach exhibits advantages in correcting OCR-generated texts, two problems emerge from the existing models:
First, some models~\cite{Lund:2009:IOC:1555400.1555437,6065414,doi:10.1117/12.2006228,doi:10.1117/12.2042502,7490117} limit candidate suggestions from the recognition output of OCR engines.
The errors unrecognized by all OCR engines are thus unable to be corrected.
This issue can be problematic especially when original input suffers from degradation, for example, historical documents~\cite{DBLP:journals/tip/NtirogiannisGP13}.
Secondly, another class of models~\cite{7490117} uses the frequencies of both candidate and related n-grams from corpus as features for training.
Although n-gram statistics is shown to be effective in correcting real-word spelling errors~\cite{Islam:2009:RSC:1699648.1699670}, training with only n-gram features does not capture the diverse nature of OCR errors and may lead to a biased model where candidates with low frequency in the corpus tend to be not selected.

In this work, we propose an OCR post-processing error correction model that leverages different features through a learning process.
Our model applies different features to avoid bias and improve the correction accuracy.
To address the limitation of candidate suggestion, we enhance the scope of the candidates of an error by considering all the words available in the vocabulary within a limited Damerau-Levenshtein distance~\cite{Damerau:1964:TCD:363958.363994} and then use features to narrow down the candidates number. 
The proposed model ranks the candidates by a regression model and shows that more than 61.5\% of the errors can be corrected on a ground truth dataset.
For 25.9\% of the uncorrected errors, our model could provide the correction in top three suggestions.

To sum up, our contributions are as follows:
\begin{itemize}
  \item We propose an OCR post-processing model which integrates OCR-specific features in a regression approach.
  The evaluation result shows that the proposed model is capable of providing high quality candidates in the top suggested list.
  \item We make available a ground truth OCR-error dataset, which is generated from a book in Biodiversity Heritage Library.
  This dataset lists the mappings from OCR-generated errors to their intended representations, which can be used directly for benchmark testing.
\end{itemize}

\section{Related Works}
\label{sec:Related-Works}


The literature of OCR post-processing research exhibits a rich family of models for correcting OCR-generated errors.
The post-processing model is an integrated system, which detects and corrects misspellings of both non-word and real-word in the OCR-generated text.

Some studies view the post-processing as the initial step in a correction pipeline and involve continuous human intervention afterwards~\cite{doi:10.1117/12.171114,Taghva:2001:OCRSpell,Muhlberger:2014:UCO:2595188.2595212}.
These models are designed to reduce the human effort in correcting errors manually.
\newcite{doi:10.1117/12.171114} integrate dictionaries and heuristics to correct as many OCR errors as possible before these are given to human correctors. 
In their future work, \newcite{Taghva:2001:OCRSpell}, record the previous human corrections to update the underlying Bayesian model for automatic correction. 
As an extreme case, \newcite{Muhlberger:2014:UCO:2595188.2595212} build a full-text search tool to retrieve all occurrences of original images given a text query, which fully relies on the user to validate and correct the errors.

One direction of work ensembles outputs from multiple OCR engines for the same input and selects the best word recognition as the final output~\cite{klein2002voting,Cecotti:2005:HOC:1106779.1107070,Lund:2009:IOC:1555400.1555437,6065414,doi:10.1117/12.2006228,doi:10.1117/12.2042502}.
\newcite{klein2002voting} show that combining complementary result from different OCR models leads to a better output.
\newcite{6065414} demonstrate that the overall error rate decreases with the addition of different OCR models, regardless of the performance of each added model.
\newcite{doi:10.1117/12.2006228} use machine learning techniques to select the best word recognitions among different OCR outputs.
\newcite{doi:10.1117/12.2042502} apply both OCR recognition votes and lexical features to train a Conditional Random Field model and evaluate the test set in a different domain.
While such models have proved useful, they select words only among OCR model recognitions and are blind to other candidates.
Besides, they require the presence of the original OCR input and effort of multiple OCR processing.

Another class of post-processing models abstracts from OCR engines and leverages statistics from external resources~\cite{DBLP:journals/ccsecis/BassilA12,DBLP:journals/corr/abs-1204-5852,7490117}.
\newcite{7490117} use three n-gram statistical features extracted from three million documents to train a linear regressor for candidate ranking.
\newcite{DBLP:journals/ccsecis/BassilA12} make use of the frequencies in the Google Web 1T n-gram corpus~\cite{googleWeb1T} for candidate suggestion and ranking.
Candidates suggested from these models are not restricted to exist in OCR recognitions.
However, existing methods make use of solely n-gram frequencies without knowing the characteristics of OCR errors and are, thus, bias to select common words from the n-gram corpus.

\section{Characteristics of OCR Errors}
\label{sec:characteristics-of-OCR-Errors}

The word error rate of OCR engines, in practice, is in the range of 7-16\%~\cite{Santos:1992:OHR,Jones:1991:IMK}, which is significantly higher than the 1.5-2.5\% for Handwriting~\cite{Wing:1980:SEH,Mitton:1987:SCC:33059.33067} and the 0.2-0.05\% for the edited newswire~\cite{Pollock:1984:ASC:358027.358048,Church:1991:PSS}.
OCR-generated errors tend to have some distinct characteristics, which require different techniques than spell correction:

\paragraph{Complex non-standard edits}
The human-generated misspellings are character-level edits, which can be categorized into one of the following four standard types: \textit{insertion}, \textit{deletion}, \textit{substitution}, and \textit{transposition}.
The majority of spell correction errors, roughly 80\%, is single edit from the intended word~\cite{Damerau:1964:TCD:363958.363994} and tend to be within one length difference~\cite{Kukich:1992:TAC:146370.146380}.
However, a significant fraction of OCR-generated errors are not one-to-one character-level edit (e.g., \textit{ri} $\rightarrow$ \textit{n} or \textit{m} $\rightarrow$ \textit{iii})~\cite{Jones:1991:IMK}.

\paragraph{Multi-factor error generation}
OCR errors are generated in different processing steps due to various factors.
Taghva and Stofsky~\cite{Taghva:2001:OCRSpell} trace the errors associated with the primary OCR steps involved in the conversion process:
(1) \textit{scanning error} caused by the low paper/print quality of the original document or the pool condition of the scanning equipment.
(2) \textit{zoning error} caused by incorrect decolumnization or complex page layout.
(3) \textit{segmentation error} caused by the broken characters, overlapping characters, and nonstandard fonts in the document.
(4) \textit{classification error} caused by the incorrect mapping from segmented pixels to a single character.

\paragraph{Multi-source dependent}
The characteristics of OCR-generated errors vary according to not only human reasons (e.g., publishers or authors) but also non-human causes (e.g., text font or input quality)~\cite{Jones:1991:IMK}.
These are especially sensitive between OCR engines.
Because different OCR engines use different techniques and features for recognition leads to a different confusion probability distribution~\cite{Kukich:1992:TAC:146370.146380}.


\section{Proposed Model}
\label{sec:Proposed-Model}

In this section, we describe in detail the processing steps of the proposed model.
We use external resources during the correction process including lexicons and a word n-gram corpus.
A lexicon\footnote{The term ``lexicon'' is usually used interchangeably with ``dictionary'' and ``word list'' in the literature.} is a list of unique words and word n-gram corpus refers to a list of n-grams (i.e., $n$ consecutive words) with observed frequency counts.
Some examples of word n-gram corpus are Google Book n-gram~\cite{Michel176} and Google Web 1T 5-gram corpus~\cite{googleWeb1T}.

To annotate, we denote English strings with text font (e.g., w$_c$, s),
vectors with bold lowercase (e.g., $\mathbf{x}$),
sets with cursive uppercase (e.g., $\mathcal{C}, \mathcal{E}$),
scalar with lower-case English or Greek characters (e.g., $y, \alpha$),
and functions followed by a bracket (e.g., $dist(*)$, $score(*)$).
The size of a collection is represented as $|| * ||$ (e.g., $|| \mathcal{D} ||$).

\subsection{Error Detection}


Error detection step identifies errors in the tokenized text, which is the first step in the correction procedure.
Since a correct word will not proceed to the further correction steps, we want to set a weak detection restriction to filter only highly confident words.
We rely on the n-gram frequency to determine the correctness of a word.
A word is detected as an error if any one of the following conditions does not fulfill.

\begin{figure}[!t]
\centering
\includegraphics[width=\columnwidth]{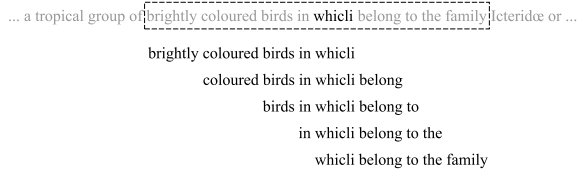}
\caption{Context collection example for token ``whicli'' in an OCR-generated text.
    Using a sliding window of size five, we are able to construct five 5-gram contexts for ``whicli''.}
\label{fig:context-example}
\end{figure}

\begin{itemize}
  \item
    Consider a common word is less likely to be an error word, the 1-gram frequency of a word should be greater than a frequency threshold.
    The frequency threshold varies with different word length.
  \item 
    A word is likely to be correct if this word with its context occurs in other places.
    We use a sliding window to construct n-gram contexts for a word.
    The frequency of one of the context in the n-gram corpus should be greater than a frequency threshold.
\end{itemize}

\subsection{Candidate Search}
\label{sec:Candidate-Search}

We select a candidate set for each error, which contains  all the words in the vocabulary within a limited number of character modifications.
To be specific, let $\Sigma$ be the symbol set, $\mathcal{L} \in \Sigma^*$ be a language lexicon.
The candidate set for a detected error $w_e$ is:
\begin{align}
  \{~w_c~ | w_c \in \mathcal{L}, dist(w_c, w_e) \le \delta \},
  \label{eq:candidate}
\end{align}
where $dist(*)$ is the minimum edit distance and $\delta$ is a distance threshold.
Damerau-Levenshtein distance~\cite{Damerau:1964:TCD:363958.363994}, which is used in most of the spell correction models for locating the candidates, considers all four character-level editing types (mentioned in Section~\ref{sec:characteristics-of-OCR-Errors}). 
Since transposition errors are common in human-generated text but rarely occur in the OCR-generated text, we apply Levenshtein distance~\cite{Levenshtein:1966:BCC}, which uses a simpler operation set without transposition.

\subsection{Feature Scoring}

We score each error candidate by features.
In this section, we discuss the contribution of the features in candidate estimation and describe the scoring measures applied in our model.

\paragraph{Levenshtein edit distance}
Minimum edit distance is a fundamental technique in quantifying the difference between two strings in spell correction.
Given two string $\text{s}_1$ and $\text{s}_2$ on alphabet $\Sigma$, the edit distance $dist(\text{s}_1, \text{s}_2)$ is the minimum number of edit operations required to transform from $\text{s}_1$ into $\text{s}_2$~\cite{Wagner:1974:ONC:360980.360995}.
An edit operation is a character-level modification in $\Sigma$.
We use Levenshtein edit distance for the same reason as previously described in Section~\ref{sec:Candidate-Search}.
The score function is as follows:
\begin{align}
  score(\text{w}_c, \text{w}_e) = 1 - \frac{dist(\text{w}_c, \text{w}_e)}{\delta + 1}
\end{align}

\paragraph{String similarity}
Longest common subsequence~\cite{Allison:1986:BLA:8871.8877} (LCS) is an alternative approach than edit distance in matching similar strings.
There are variations of LCS:
\textit{Normalized Longest Common Subsequence (NLCS)}, which take into account the length of both the shorter and the longer string for normalization.
\begin{align}
  nlcs(\text{w}_c, \text{w}_e) = \frac{2 \cdot len(lcs(\text{w}_c, \text{w}_e))^2}{len(\text{w}_c) + len(\text{w}_e)}.
\end{align}
\textit{Normalized Maximal Consecutive Longest Common Subsequence (MCLCS)}, which limits the common subsequence to be consecutive.
There are three types of modifications with different additional conditions:
\emph{NLCS}$_1$ and \emph{NLCS}$_n$ use the subsequences starting at the first and the $n$-th character, respectively;
\emph{NLCS}$_z$ takes the subsequences ending at the last character.
They apply the same normalization as \emph{NLCS}.
\begin{align}
  & nmnlcs_1(\text{w}_c, \text{w}_e) = \frac{2 \cdot len(mclcs_1(\text{w}_c, \text{w}_e))^2}{len(\text{w}_c) + len(\text{w}_e)} \\
  & nmnlcs_n(\text{w}_c, \text{w}_e) = \frac{2 \cdot len(mclcs_n(\text{w}_c, \text{w}_e))^2}{len(\text{w}_c) + len(\text{w}_e)} \\
  & nmnlcs_z(\text{w}_c, \text{w}_e) = \frac{2 \cdot len(mclcs_z(\text{w}_c, \text{w}_e))^2}{len(\text{w}_c) + len(\text{w}_e)}.
\end{align}
We apply the measure proposed in~\cite{Islam:2009:RSC:1645953.1646205} for scoring, which takes the weighted sum of the above LCS variations:
\begin{align}
  &score(\text{w}_c, \text{w}_e) \nonumber \\
  &\ = \alpha_1 \cdot nlcs(\text{w}_c, \text{w}_e) + \alpha_2 \cdot nmnlcs_1(\text{w}_c, \text{w}_e) \nonumber \\
  &\ + \alpha_3 \cdot nmnlcs_n(\text{w}_c, \text{w}_e) + \alpha_4 \cdot nmnlcs_z(\text{w}_c, \text{w}_e).
\end{align}

\paragraph{Language popularity}
Using a language lexicon is a common approach to detect the non-word tokens, where non-existing tokens are detected as true errors.
Let $\text{w}_c$ be the candidate string, $\mathcal{C}$ be the set of all error candidates, and $freq_1(\cdot)$ be the unigram frequency.
The candidate confidence is the unigram popularity given by:
\begin{align}
  score(\text{w}_c, \text{w}_e) = \frac{freq_1(\text{w}_c)}{\max_{\text{w}_c' \in C} freq_1(\text{w}_c')}.
\end{align}

\paragraph{Lexicon existance}
Besides English lexicon, we can use different lexicons to detect the existence of the token in different subjects. 
It identifies additional lexical features.
For example, we may use a domain specific lexicon to capture terminologies, which is especially useful for input text from the same domain.
The candidate selection is the same as English lexicon, but the candidate score is a boolean value that indicates the detection result.
\begin{align}
  score(\text{w}_c, \text{w}_e) =
    \begin{cases}
      1 & \text{if $\text{w}_c$ exists in the lexicon} \\
      0 & \text{otherwise} \\
    \end{cases}
\end{align}

\paragraph{Exact-context popularity}
An appropriate correction candidate should be coherent in context.
Using word n-gram for context analysis is a broadly researched approach in correcting real-word errors~\cite{Islam:2009:RSC:1645953.1646205}. 
Given an error word $\text{w}_e$ in a text, we have its n-gram contexts $\mathcal{G}$ constructed using a sliding window (see Figure~\ref{fig:context-example}).
To score a candidate $\text{w}_c$ of this error, we first substitute the error word from each of its n-gram contexts by such candidate and create a new set of contexts $\mathcal{G}_c$.
Let $\mathcal{C}$ be all candidates suggested for $\text{w}_e$, and $freq_n(\cdot)$ be the n-gram frequency, which gives 0 to a non-existing n-gram.
The score function is given as:
\begin{align}
  score(\text{w}_c, \text{w}_e) = \frac{\sum_{\mathbf{c} \in \mathcal{G}_c} freq_n(\mathbf{c})}
  {\max_{\text{w}'_c \in \mathcal{C}} \{ \sum_{\mathbf{c}' \in \mathcal{G}'_c} freq_n(\mathbf{c}') \}}
  \label{eq:exact-score}
\end{align}

\paragraph{Relaxed-context popularity}
A context with longer n-gram size defines a more specific use case for a given word, where its existence in the corpus shows higher confidence for a candidate.
In general, an n-gram corpus has limited coverage for all possible n-grams in the language, especially for the emerging words in the language.
Candidates of a rare word can barely be suggested from its contexts because of the limited coverage in the n-gram corpus.
We deal with such issue by relaxing the context matching condition to allow one mismatching context word.
For example, in Figure~\ref{fig:context-example}, we consider only the first 5-gram context given ``which'' be the candidate. 
We need the frequency of ``brightly coloured birds in which'' for computing exact context popularity.
As for the relaxed context popularity, we need to sum up the frequencies of four types of 5-grams: ``* coloured birds in which'', ``brightly * birds in which'', ``brightly coloured * in which'', and ``brightly coloured birds * which'', where * matches any valid unigram.
The scoring function is the same as the exact context matching (Eq.~\ref{eq:exact-score}), except for the candidate set and the context set are larger in the relaxed case.

\subsection{Candidate Ranking}

We formulate the confidence prediction task as a regression problem.
Given candidate feature scores, we predict the confidence of each candidate being a correction for the error word.
The confidence is used for ranking among candidates of one error.

To train a regressor for correction, we label candidate features with 1 if a candidate is the intended correction, or 0 otherwise. 
The training data contains candidates from different errors, and there are more candidates labeled 0 than 1.
To deal with the unbalanced nature of the candidates, we weight the samples when computing the training loss
\begin{align}
  loss(\mathcal{D}) = \sum_{e \in \mathcal{E}}\sum_{c \in {\mathcal{C}^\mathcal{F}_e}} w_{c} \cdot loss(\mathbf{x}_c, y_c).
\end{align}
We count the number of samples with label 1 and 0, respectively.
Then, we use the ratio to weight for samples labeled 1, and 1 for samples labeled 0.

Experimentally, we apply a AdaBoost.R2~\cite{Freund:1997:DGO:261540.261549} model on top of decision trees with linear loss function.

\section{Evaluation}
\label{sec:Evaluation}

To better describe the error sample, we use the annotation $<$w$_t$$\rightarrow$w$_e$$>$ to represent the intended word w$_t$ being recognized as the error word w$_e$.

\subsection{Experimental Dataset}

\begin{table*}[!t]
  \small
  \centering
  \caption{The Levenshtein edit distance distribution of error in the experimental dataset.}
  \vspace{1em}
  \label{tab:error-distribute}
  \begin{tabular}{c r r l l}
    \toprule
    \multirow{2}{2.5cm}{\centering Levenshtein edit distance} & \multicolumn{2}{c}{Error Statistics} & \multicolumn{2}{c}{Sample Error} \\
    \cmidrule(lr){2-3}
    \cmidrule(lr){4-5}
    & Number & Percentage                  & Intended Word & Error Word \\
    \midrule
    1        & 669  & 24.60\% & galbula   & ga/bula \\[0.4em]
    2        & 1353 & 49.76\% & yellowish & j\string^ellowish \\[0.4em]
    3        & 296  & 10.89\% & bents     & Ijcnts \\[0.4em]
    4        & 163  &  5.99\% & my & ni\}' \\[0.4em]
    5        & 82   &  3.02\% & Lanius & Lioiiits \\[0.4em]
    6        & 52   &  1.91\% & minor &  )iii$>$iof \\[0.4em]
    7        & 29   &  1.07\% & garrulus & f;ay$>$///us \\[0.4em]
    8        & 19   &  0.70\% & curvirostra & iUi'7'iyosira \\[0.4em]
    9        & 9   &  0.33\% & Nucifraga & Aiiii/rut\string^d \\[0.4em]
    $\ge 10$ & 26   &  0.96\% & pomeranus & poiiui-iVtiis \\
    \midrule
    total & 2698 & 100\% & \\
    \bottomrule
  \end{tabular}
\end{table*}


We made available a dataset with 2728 OCR-generated errors along with the ground truth and OCR text for benchmark testing.
The OCR text was generated from the book titled ``Birds of Great Britain and Ireland''~\cite{bhl35947} and made it publicly available by the Biodiversity Heritage Library (BHL) for Europe\footnote{\url{http://www.biodiversitylibrary.org/item/35947#page/13/mode/1up}}.
The ground truth text is based on an improved OCR output\footnote{http://www.bhle.eu/en/results-of-the-collaboration-of-bhl-europe-and-impact} and adjusted manually to match with the original content of the whole book.

This source image data of the book contains 460 page-separated files, where the the main content is included in 211 pages.
This book combines different font types and layouts in main text, which leads erroneous OCR results.
There are 2698 mismatching words between the ground truth text with the BHL digital OCR-text, which are used as the ground truth errors.
The ground truth text contains 84492 non-punctuation words.
Thus, the OCR error rate of the evaluation dataset is 3.22\%, where some errors are complex regarding edit distance, shown in Table.~\ref{tab:error-distribute}.
Other challenges include terminologies in multilingual (e.g., Turdid{\ae}, Fringillid{\ae}) and meanless words (e.g., bird-sound simulation: ``cir-ir-ir-ir-re'', ``vee-o''), which may not be handled using the standard techniques.

\subsection{Evaluation Setup}

\newcite{Walker:2014:TUM} claim that the lexicon from a published dictionary has limited coverage on newswire vocabulary, and vice versa.
Thus, we construct a language lexicon with unigrams in the Google Web 1T n-gram corpus\footnote{https://catalog.ldc.upenn.edu/LDC2006T13}.
This corpus contains the frequencies of unigrams (single words) to five-grams, which is generated from approximately 1 trillion word tokens extracted from publicly accessible Web pages.
Its unigram corpus is filtered with the frequency no less than 200.
We use five-grams in Google Web 1T corpus for exact and relaxed context matching.

For lexicon existence feature, we use three lexicons to build two features instances:
(1) \emph{Wikipedia entities} extracted from article names in Wikipedia.
    This feature gives credit to common terminologies.
(2) \emph{Biodiversity terminologies} collected from biodiversity digital library to capture the domain specific terms, which may not be contained in Wikipedia.

The proposed model receives OCR-generated plain text as input.
We apply the Penn Treebank tokenization with the additional rules from Google\footnote{https://catalog.ldc.upenn.edu/docs/LDC2006T13/readme.txt} to tokenize the input text.
This tokenization method is consistent with the Google Web 1T n-gram corpus.
The frequency and existence of rarely hyphenated words can be poorly estimated using external resources.
Thus we split the hyphenated word by the internal hyphen.

Experimentally, we filter tokens of the following types after tokenization:
(1) \emph{punctuations};
(2) \emph{numeric tokens}, which contains only numeric characters (i.e., 0-9);
(3) \emph{common English words}. We apply a lexicon of frequent English words for filtering.
The accuracy of the system will increase with more relaxed filtering conditions on English words, for example, filtering only English stop words or even no filtering, but the computation time increases as the trade-off.
Similarly for reducing the candidate detection time in Eq.~\ref{eq:candidate}, we set the maximum Levenshtein distance $\delta$ for candidate search to be 3.

\subsection{Detection Evaluation}

\newcommand\MyBox[1]{
  \fbox{\lower0.5cm
    \vbox to 1.25cm{\vfil
      \hbox to 1.25cm{\hfil\parbox{0.9cm}{\centering #1}\hfil}
      \vfil}%
  }%
}

\begin{table}[!t]
  \caption{Confusion matrix for error detection}
  \label{tab:detection-confusion}
  \vspace{0.5em}
  \small
  \centering
  \renewcommand\arraystretch{1.5}
  \setlength\tabcolsep{0pt}
  \begin{tabular}{c >{\bfseries}r @{\hspace{0.7em}}c @{\hspace{0.4em}}c @{\hspace{0.7em}}l}
    \multirow{10}{*}{\rotatebox{90}{\lapbox{-1em}{\bfseries\centering Actual Correctness}}} & 
      & \multicolumn{2}{c}{\bfseries Model Detection} & \\
    & & \bfseries Error & \bfseries Correct & \bfseries total \\
    & \rotatebox{90}{\lapbox{-1em}{Error}}   & \MyBox{2457} & \MyBox{241}   & 2698  \\[2.0em]
    & \rotatebox{90}{\lapbox{-1em}{Correct}} & \MyBox{1273} & \MyBox{80523} & 81794 \\
    & total & 3730 & 80764 &
  \end{tabular}
\end{table}

We evaluate error detection as a recall oriented task, which focus on finding all possible errors.
In all error correction techniques, an undetected error will not get into the correction phase.

We report the confusion matrix for error detection in Table~\ref{tab:detection-confusion}.
The proposed model achieves 91.07\% detection recall.
There are considerable number of ture-positive errors, which are correct words but detected as errors.
When using this type of errors for training or testing, we use the word itself as the intended word for each error.
The correction results regarding to all types of errors are reported in Section~\ref{sec:Correction-Evaluation}.

For tokenizing the noisy text, any tokenization approach is inevitably involved in the common word boundary problem~\cite{Kukich:1992:TAC:146370.146380}, the correct boundary of the errors are not properly identified, in both human-generated~\cite{Kukich:1992:SCT:129875.129882} and OCR-generated text~\cite{Jones:1991:IMK}.
Such problem can be caused by the splitting (e.g., $<$\textit{spend}$\rightarrow$\textit{sp end}$>$) and merging (e.g., $<$\textit{in form}$\rightarrow$\textit{infrom}$>$) mistakes.
It is especially problematic in OCR-generated text, where words containing characters are recognized as punctuation and are thus splitted by the tokenization heuristics.
Most error detection and correction techniques define token as character sequence separated by white space characters (e.g., blanks, tabs, carriage returns, etc.)~\cite{Kukich:1992:TAC:146370.146380}, which do not split the error token by punctuations.
However, this approach cannot distinguish between true punctuation and misrecognized trailing punctuation (e.g., $<$\textit{family}$\rightarrow$\textit{famil\}\string^}$>$).


\begin{table}[!t]
  \centering
  \caption{The number of detected errors and recall of bounded and unbounded detections}
  \label{tab:detction-accuracy}
  \vspace{0.5em}
  \small
  \begin{tabular}{c r r r}
    \toprule
    Detection Category & Number  & Recall
    \\
    \midrule                                                               
    Bounded        & 1995   & 73.94\%  \\[0.4em]
    Unbounded      &  462   & 17.12\%  \\
    \midrule                           
    Total (True-Positive)  & 2457   & 91.07\%  \\
    \bottomrule
  \end{tabular}
\end{table}

An error may be ``partially'' detected if an overlapped but non-identical character sequence is treated as an error.
We call this ``partially'' detected case as a success \emph{unbounded} detection, where the correct recognition of the character sequence as success \emph{bounded} detection.
Unbounded detection can potentially be corrected, but it has inaccurate features scores that will influence the correction accuracy.
For example, if an error $<$\emph{spend}$\rightarrow$\emph{sp end}$>$ is unbounded and detected as \emph{end}, \emph{sp} will exists in the context and candidate edit distance will be computed with \emph{end} instead of \emph{sp end}.
In addition, there may exist multiple unbounded errors detected for one ground truth error, because of the splitting mistakes.
For every ground truth error, we count at most one successful unbounded detection.
Our model achieves the $73.51\%$ bounded detection recall and $90.51\%$ total detection recall (i.e., sum of bounded and unbounded detection) shown in Table~\ref{tab:detction-accuracy}.


\subsection{Correction Evaluation}
\label{sec:Correction-Evaluation}

We take the following steps to build a training dataset:
First, we construct a candidate set for each error containing top 10 candidates scored by each feature.
Then, we select a subset of errors, whose intended word exists in the candidate set.
Finally, we randomly select 80\% errors and use their candidates sets for training.

We train multiple AdaBoost regressors with different settings and apply 10-fold cross-validation to select the best setting for evaluating the rest errors.
We report the correction results regarding different error categories in Table~\ref{tab:corr-result}.
\emph{P@n} represents precision at top $n$ candidate suggestions, which calculate the ratio of the existence of intended words in top $n$ candidates.
The proposed model rank the candidates by a regression model and show that more than 61.5\% of the errors can be corrected.
For 25.9\% of the uncorrected errors, our model could provide the correction in the top three suggestions.

\begin{table}[!t]
  \centering
  \caption{The percentage of errors, where correction exists among top 1, 3, 5, and 10 candidates suggested by the proposed model.}
  \label{tab:corr-result}
  \vspace{0.5em}
  \small
  \begin{tabular}{l c c c c}
    \toprule
    Error Categories & P@1 & P@3 & P@5 & P@10 \\
    \midrule
    Bounded        & 0.6369 & 0.7710 & 0.8028 & 0.8405 \\
    Unbounded      & 0.5637 & 0.6417 & 0.6620 & 0.6823 \\
    \midrule         
    True-Positive  & 0.6095 & 0.7025 & 0.7238 & 0.7604 \\
    False-Positive & 0.6971 & 0.7145 & 0.7738 & 0.7942 \\
    \midrule         
    Total          & 0.6150 & 0.7145 & 0.7378 & 0.7662 \\
    \bottomrule
  \end{tabular}
\end{table}

\section{Discussion}
\label{sec:Discussion}

\subsection{Selected Features}

\begin{table}[!t]
  \centering
  \caption{The number and the percentage of errors, where correction exists among the top 10 candidates of any applied feature.}
  \label{tab:search-statistics}
  \vspace{0.5em}
  \small
  \begin{tabular}{c r R{1.45cm} R{1.8cm}}
    \toprule
    \raisebox{-0.3em}{\multirow{2}{1.65cm}{\centering Detection Category}} &
    \multicolumn{3}{c}{Correct Candidates}
    \\
    \cmidrule(lr){2-4}
    & Number  & Percentage Among All & Percentage in Search Scope
    \\
    \midrule                                                               
    Bounded        & 1540    & 77.19\%    & 84.71\%     \\[0.4em]
    Unbounded      & 108     & 23.38\%    & 47.58\%     \\
    \midrule                                           
    True-Positive  & 1648    & 67.07\%    & 78.66\%     \\[0.4em]
    False-Positive & 1273    & 100.00\%   & 100.00\%     \\
    \midrule                                           
    Total          & 2627    & 66.15\%    & 78.00\%     \\
    \bottomrule
  \end{tabular}
\end{table}

\pgfplotstableread{
Label           f1    f2    f3    f4    f5    f6
Exact           93	171	102	78	100	67
Relax           68	206	148	146	188	67
Lexcon          12	39	95	256	99	67
Lang            44	96	295	339	187	67
Sim             2	34	148	258	179	67
Distance        17	74	214	299	187	67
}\searchboundthree
\pgfplotstableread{
Label           f1    f2    f3    f4    f5    f6
Exact           3	6	2	0	4	0
Relax           9	10	2	3	5	0
Lexcon          0	20	11	7	1	0
Lang            15	27	20	10	5	0
Sim             0	2	15	10	5	0
Distance        1	3	19	10	5	0
}\searchunboundthree
\pgfplotstableread{
Label           f1    f2    f3    f4    f5    f6
Exact           0	0	0	0	0	0
Relax           0	0	0	2	10	0
Lexcon          0	0	475	328	10	0
Lang            0	1	31	326	10	0
Sim             0	132	506	328	10	0
Distance        2	133	506	328	10	0
}\searchfnthree

\pgfplotstableread{
Label           f1    f2    f3    f4    f5    f6
Exact           85	189	68	72	67	151
Relax           95	217	109	115	247	151
Lexcon          16	44	56	447	206	151
Lang            16	59	159	485	247	151
Sim             0	6	44	403	243	151
Distance        0	23	86	426	225	151
}\searchboundten
\pgfplotstableread{
Label           f1    f2    f3    f4    f5    f6
Exact           2	7	1	1	0	5
Relax           11	7	3	21	8	5
Lexcon          2	4	4	45	8	5
Lang            2	7	14	49	8	5
Sim             0	2	11	48	8	5
Distance        0	1	12	32	8	5
}\searchunboundten
\pgfplotstableread{
Label           f1    f2    f3    f4    f5    f6
Exact           0	0	0	0	0	31
Relax           0	0	0	3	93	31
Lexcon          0	0	51	642	93	31
Lang            0	2	157	645	93	31
Sim             0	0	208	645	93	31
Distance        0	2	208	645	93	31
}\searchfnten

\begin{figure*}[ht!]
  \begin{center}
  \begin{tikzpicture}
    \begin{groupplot}[
        group style={
            group size=3 by 2,
            columns=3,
            vertical sep=7pt,
            horizontal sep=7pt
        },
        width=\textwidth/3,
        xbar stacked,                       
        xmin=0,                             
        every extra x tick/.style={xmajorgrids=true, tick label style={yshift=18pt, xshift=10pt}},
      ]
      \nextgroupplot[title={\textbf{Bounded}}, ytick=data, yticklabels from table={\searchboundthree}{Label},
        ylabel={Top 3},
        ylabel style={yshift=10pt, font=\bfseries},
        xmax=1818,
        xticklabels={},
        extra x ticks={1462},
        extra x tick style={tick label style={xshift=-25pt}, xticklabel=\pgfmathprintnumber{\tick}}
      ]
        \addplot [fill=white]               table [x=f6, y expr=\coordindex] {\searchboundthree};
        \addplot [pattern=dots]             table [x=f5, y expr=\coordindex] {\searchboundthree};
        \addplot [pattern=north west lines] table [x=f4, y expr=\coordindex] {\searchboundthree};
        \addplot [pattern=crosshatch]       table [x=f3, y expr=\coordindex] {\searchboundthree};
        \addplot [pattern=crosshatch, postaction={pattern=grid}]       table [x=f2, y expr=\coordindex] {\searchboundthree};
        \addplot [fill=black]               table [x=f1, y expr=\coordindex] {\searchboundthree};
      \nextgroupplot[title={\textbf{Unbounded}}, yticklabels={,,},
        xmax=277,
        xticklabels={},
        extra x ticks={99},
        extra x tick style={xticklabel=\pgfmathprintnumber{\tick}}
      ]
        \addplot [fill=white]               table [x=f6, y expr=\coordindex] {\searchunboundthree};
        \addplot [pattern=dots]             table [x=f5, y expr=\coordindex] {\searchunboundthree};
        \addplot [pattern=north west lines] table [x=f4, y expr=\coordindex] {\searchunboundthree};
        \addplot [pattern=crosshatch]       table [x=f3, y expr=\coordindex] {\searchunboundthree};
        \addplot [pattern=crosshatch, postaction={pattern=grid}]       table [x=f2, y expr=\coordindex] {\searchunboundthree};
        \addplot [fill=black]               table [x=f1, y expr=\coordindex] {\searchunboundthree};
      \nextgroupplot[title={\textbf{False-Positive}}, yticklabels={,,},
        xmax=1273,
        xticklabels={},
        extra x ticks={977},
        extra x tick style={xticklabel=\pgfmathprintnumber{\tick}}
      ]
        \addplot [fill=white]               table [x=f6, y expr=\coordindex] {\searchfnthree};
        \addplot [pattern=dots]             table [x=f5, y expr=\coordindex] {\searchfnthree};
        \addplot [pattern=north west lines] table [x=f4, y expr=\coordindex] {\searchfnthree};
        \addplot [pattern=crosshatch]       table [x=f3, y expr=\coordindex] {\searchfnthree};
        \addplot [pattern=crosshatch, postaction={pattern=grid}]       table [x=f2, y expr=\coordindex] {\searchfnthree};
        \addplot [fill=black]               table [x=f1, y expr=\coordindex] {\searchfnthree};
      \nextgroupplot[ytick=data, yticklabels from table={\searchboundten}{Label},
        ylabel={Top 10},
        ylabel style={yshift=10pt, font=\bfseries},
        xmax=1818,
        extra x ticks={1540},
        extra x tick style={tick label style={xshift=-25pt}},
      ]
        \addplot [fill=white]               table [x=f6, y expr=\coordindex] {\searchboundten};
        \addplot [pattern=dots]             table [x=f5, y expr=\coordindex] {\searchboundten};
        \addplot [pattern=north west lines] table [x=f4, y expr=\coordindex] {\searchboundten};
        \addplot [pattern=crosshatch]       table [x=f3, y expr=\coordindex] {\searchboundten};
        \addplot [pattern=crosshatch, postaction={pattern=grid}]       table [x=f2, y expr=\coordindex] {\searchboundten};
        \addplot [fill=black]               table [x=f1, y expr=\coordindex] {\searchboundten};
      \nextgroupplot[yticklabels={,,},
        xmax=277,
        extra x ticks={108},
        legend image code/.code={
            \draw[#1] (0cm,-0.1cm) rectangle (0.6cm,0.1cm);
        },
        legend style={
            at={(0.5,-0.2)},
            anchor=north,
            draw=none,
            legend columns=-1,
            /tikz/every even column/.append style={column sep=0.1cm}
        },
      ]
        \addlegendentry{\textbf{Error Corrections are located by the number of features:}} 
        \addlegendimage{empty legend}
        \addplot [fill=white]               table [x=f6, y expr=\coordindex] {\searchunboundten}; \addlegendentry{6,}
        \addplot [pattern=dots]             table [x=f5, y expr=\coordindex] {\searchunboundten}; \addlegendentry{5,}
        \addplot [pattern=north west lines] table [x=f4, y expr=\coordindex] {\searchunboundten}; \addlegendentry{4,}
        \addplot [pattern=crosshatch]       table [x=f3, y expr=\coordindex] {\searchunboundten}; \addlegendentry{3,}
        \addplot [pattern=crosshatch, postaction={pattern=grid}]       table [x=f2, y expr=\coordindex] {\searchunboundten}; \addlegendentry{2,}
        \addplot [fill=black]               table [x=f1, y expr=\coordindex] {\searchunboundten}; \addlegendentry{1}
      \nextgroupplot[yticklabels={,,},
        xmax=1273,
        extra x ticks={979}
      ]
        \addplot [fill=white]               table [x=f6, y expr=\coordindex] {\searchfnten};
        \addplot [pattern=dots]             table [x=f5, y expr=\coordindex] {\searchfnten};
        \addplot [pattern=north west lines] table [x=f4, y expr=\coordindex] {\searchfnten};
        \addplot [pattern=crosshatch]       table [x=f3, y expr=\coordindex] {\searchfnten};
        \addplot [pattern=crosshatch, postaction={pattern=grid}]       table [x=f2, y expr=\coordindex] {\searchfnten};
        \addplot [fill=black]               table [x=f1, y expr=\coordindex] {\searchfnten};
    \end{groupplot}
  \end{tikzpicture}
  \caption{The distinctiveness of features in locating error corrections. 
    A bar of a feature represents the number of error corrections located by this feature.
    The color of a bar indicates the number of features that locates these errors.
    i.e., white bar indicates a portion of error corrections located by all the features, while black bar indicates error corrections located by only one feature.
    }
  \label{fig:feature-search}
  \end{center}
\end{figure*}

We want to study the contribution of features to candidate suggestion.
We first explore how well the scoring functions could rank the intended words to the top without predicting by the regressor.
For each detected error, we construct a candidate set containing top 10 candidates scored by each feature and check whether this candidate set contains a correction.
Note that candidate search scope is limited by the number of edit distance $\delta$ (in Eq.~\ref{eq:candidate} by default), thus the intended words w$_t$ for w$_e$ cannot be found if $dist_{lev}(\text{w}_t, \text{w}_e) > \delta$.
Results are shown in Table~\ref{tab:search-statistics}.
The model could locate most of the correction in top candidates with the collaboration of all applied features.
We observe that performance varies drastically for bounded and unbounded errors, presumably because the feature score for unbounded errors is inaccurate (e.g., the split part of a splitting error is counted as the context word in context search).

To get a better intuition for the contribution of individual feature, we plot the distinctiveness of the located error corrections by each feature in Figure~\ref{fig:feature-search}. For bounded detected errors, context-based features are able to locate some distinctive corrections, which can rarely be found by other features.
In addition, Relax context popularity feature shows better coverage than exact context popularity.
On the other hand, the other four features are important for false-positive errors, where context-based features provide little help.

%
%
%
%
%
%

\subsection{Regression Model Selection}

We report candidate ranking performance of different regression models in Table~\ref{tab:regress-model-select}.
The same training and testing dataset, described in Section~\ref{sec:Correction-Evaluation}, are used for all models.

Given the upperbound of correction rate within three edit distance is 78\% (in Table~\ref{tab:search-statistics}), all regressors achieve good results.
As can be seen, ensemble methods, like Random Forest and AdaBoost, are more robust than others in suggesting appropriate candidates.


\begin{table}[!t]
  \centering
  \caption{The percentage of errors, where correction exists among top 1, 3, 5, and 10 candidates suggested by
  Support Vector regressor (SVR), Rigid Linear regressor (RL), Multiple Layer Perceptron with rectified linear unit (MLP.ReLU), Random Forest (RF), and AdaBoost.R2}
  \label{tab:regress-model-select}
  \vspace{0.5em}
  \small
  \begin{tabular}{l c c c c}
    \toprule
    \multirow{2}{*}{Regression Model} &\multicolumn{4}{c}{All Errors} \\
    \cmidrule(lr){2-5}
    & P@1 & P@3 & P@5 & P@10 \\
    \midrule
    SVR              & 0.5634 & 0.7158 & 0.7378 & 0.7612 \\
    RL               & 0.5637 & 0.6817 & 0.7020 & 0.7313 \\
    MLP.ReLU + BFGS  & 0.6095 & 0.7025 & 0.7283 & 0.7604 \\
    RF               & 0.6071 & 0.7145 & 0.7378 & 0.7516 \\
    AdaBoost.R2 + DT & 0.6150 & 0.7145 & 0.7378 & 0.7662 \\
    \bottomrule
  \end{tabular}
\end{table}

\section{Conclusion}
\label{sec:Conclusion}

We introduce a statistical learning model for correcting OCR-generated errors.
By integrating different features in a regression process, our model is able to select and rank candidates that are similar in shape to the error, suitable for the domain, and coherent to the context.
The evaluation results show that our model can correct 61.5\% of the errors and could provide a correction in top three suggestions for 25.9\% of the uncorrected errors.
That is, by suggesting three candidates for each error, our model can correct 71.5\% error cases in a theoretical correction upperbound of 78\%.


\ifx\arXiv\undefined
  \bibliography{bib/main,bib/survey,bib/spell-correct,bib/ocr-detect,bib/cor-correct,bib/others}
\else

\fi
\bibliographystyle{template/eacl2017}


\end{document}